\documentclass[sigconf]{acmart}
\pdfoutput=1
\usepackage{booktabs} 

\usepackage{amsmath}
\usepackage{amssymb}
\usepackage{booktabs} 
\usepackage{tikz}
\usepackage{xcolor}
\usepackage{multirow}
\usetikzlibrary{bayesnet}

\newcommand{\eat}[1]{}

\usepackage{dcolumn}
\newcolumntype{d}[1]{D{.}{.}{#1} }

\usepackage{listings}
\usepackage{setspace}
\definecolor{Code}{rgb}{0,0,0}
\definecolor{Decorators}{rgb}{0.5,0.5,0.5}
\definecolor{Numbers}{rgb}{0.5,0,0}
\definecolor{MatchingBrackets}{rgb}{0.25,0.5,0.5}
\definecolor{Keywords}{rgb}{0,0,1}
\definecolor{self}{rgb}{0,0,0}
\definecolor{Strings}{rgb}{0,0.63,0}
\definecolor{Comments}{rgb}{0,0.63,1}
\definecolor{Backquotes}{rgb}{0,0,0}
\definecolor{Classname}{rgb}{0,0,0}
\definecolor{FunctionName}{rgb}{0,0,0}
\definecolor{Operators}{rgb}{0,0,0}
\definecolor{Background}{rgb}{0.98,0.98,0.98}
\lstdefinelanguage{Python}{
numbers=left,
numberstyle=\footnotesize,
numbersep=1em,
xleftmargin=1em,
framextopmargin=2em,
framexbottommargin=2em,
showspaces=false,
showtabs=false,
showstringspaces=false,
frame=l,
tabsize=4,
basicstyle=\ttfamily\footnotesize\setstretch{1},
backgroundcolor=\color{Background},
commentstyle=\color{Comments}\slshape,
stringstyle=\color{Strings},
morecomment=[s][\color{Strings}]{"""}{"""},
morecomment=[s][\color{Strings}]{'''}{'''},
morekeywords={import,from,class,def,for,while,if,is,in,elif,else,not,and,or,print,break,continue,return,True,False,None,access,as,,del,except,exec,finally,global,import,lambda,pass,print,raise,try,assert},
keywordstyle={\color{Keywords}\bfseries},
morekeywords={[2]@invariant,pandas,numpy,scipy},
keywordstyle={[2]\color{Decorators}\slshape},
emph={self},
emphstyle={\color{self}\slshape},
}

\copyrightyear{2019}
\acmYear{2019}
\setcopyright{iw3c2w3}
\acmConference[WWW '19]{Proceedings of the 2019 World Wide Web Conference}{May 13--17, 2019}{San Francisco, CA, USA}
\acmBooktitle{Proceedings of the 2019 World Wide Web Conference (WWW '19), May 13--17, 2019, San Francisco, CA, USA}
\acmPrice{}
\acmDOI{10.1145/3308558.3313459}
\acmISBN{978-1-4503-6674-8/19/05}

\usepackage{balance}

\begin{document}

\title{Truth Inference at Scale: A Bayesian Model for Adjudicating Highly Redundant Crowd Annotations}

\author{Yuan Li}
\affiliation{%
  \institution{University of Melbourne}
  \city{Melbourne}
  \state{Australia}
}
\email{yuanl4@student.unimelb.edu.au}

\author{Benjamin I. P. Rubinstein}
\affiliation{%
  \institution{University of Melbourne}
  \city{Melbourne}
  \state{Australia}
}
\email{brubinstein@unimelb.edu.au}

\author{Trevor Cohn}
\affiliation{%
  \institution{University of Melbourne}
  \city{Melbourne}
  \state{Australia}
}
\email{tcohn@unimelb.edu.au}


\begin{abstract}
Crowd-sourcing is a cheap and popular means of creating training and evaluation datasets for machine learning, however it poses the problem of `truth inference', as individual workers cannot be wholly trusted to provide reliable annotations.
Research into models of annotation aggregation attempts to infer a latent `true' annotation, which has been shown to improve the utility of crowd-sourced data.
However, existing techniques beat simple baselines only in low redundancy settings, where the number of annotations per instance is low ($\le 3$), or in situations where workers are unreliable and produce low quality annotations (e.g., through spamming, random, or adversarial behaviours.)
As we show, datasets produced by crowd-sourcing are often not of this type: the data is highly redundantly annotated ($\ge 5$ annotations per instance), and the vast majority of workers produce high quality outputs.
In these settings, the majority vote heuristic performs very well, and most truth inference models underperform this simple baseline.
We propose a novel technique, based on a Bayesian graphical model with conjugate priors, and simple iterative expectation-maximisation inference.
Our technique produces competitive performance to the state-of-the-art benchmark methods, and is the only method that significantly outperforms the majority vote heuristic at one-sided level $0.025$, shown by significance tests.
Moreover, our technique is simple, is implemented in only $50$ lines of code, and trains in seconds.
\footnote{The code and the datasets are made available at https://github.com/yuan-li/truth-inference-at-scale.}


\end{abstract}

%
%
%


\maketitle

\section{Introduction}
Large volumes of labelled data are crucial for machine learning, retrieval and search, both for training supervised methods and for evaluation. Ideally, ground truth labels are collected from experts, but such sources are often too slow or costly, motivating the use of data annotation by crowdsourcing~\citep{howe2008crowdsourcing,callison2010creating}.
This has been shown to be highly effective for tasks which are difficult for machines to automate but relatively easy for humans, such as document categorization, image annotation, and sentiment analysis.
Using non-expert labels for training machine learning algorithms can be as effective as using gold standard annotations from experts \citep{snow2008cheap,novotney2010cheap}, despite being much cheaper and faster to obtain.

Unfortunately, crowdsourced labels are often noisy or biased,
which may be due to workers' lack of expertise, workers cheating or otherwise not performing the annotation task correctly~\citep{difallah2012mechanical}. This problem can be alleviated by filtering workers based on their profile and working history. However, this problem is not fully solved by simple quality control mechanisms currently provided by crowdsourcing platforms; consequently, most crowd data studies collect several labels for each item from different workers, which are aggregated post-hoc to infer an accurate consensus annotation.

Both the average number of labels per item (\#labels/\#items), and the quality of worker labels, are important factors in the performance of aggregation methods. Low-redundancy (e.g. $\textrm{\#labels/\#items}$ $\le 3$) and/or low-quality scenarios (e.g. there are adversarial workers) are challenging and have attracted much attention. Such scenarios require methods to make full use of a few available labels by being able to interpret labels from different workers.
Models which incorporate per-worker confusion matrices \citep{dawid1979maximum, raykar2010learning,kim2012bayesian,venanzi2014community}
are flexible enough to capture different types of worker behaviour, and have shown to be effective in low-redundancy and/or low-quality scenarios.

However, the simplest aggregation technique, majority voting (MV), still predominates in practice.
MV is an ensemble where every worker is given equal vote, based on the assumption that with increasing redundancy, the most common label will converge to the true label.
Despite its simplicity, on many real-world crowdsourced datasets, MV outperforms most models of truth inference, as we show in our experimental evaluation, and the best method from prior work only outperforms MV by a small margin.
To explain this observation, and the continued popularity of MV, we identify several misalignments between the experimental setup in previous papers and real-world scenarios:
\begin{description}
\item[Redundancy \& worker quality.] In practice \#labels/\#items is usually $5$+ to ensure sufficient redundancy, and crowdsourcing platforms provide mechanisms to select high quality workers (e.g. in the collection of question answering dataset SQuAD~\citep{rajpurkar2016squad}, workers were filtered based on the amount and acceptance rate of their prior work history). In such settings, MV is very competitive. Worryingly, our results show that confusion-matrix-based methods are inferior to MV in terms of mean accuracy across datasets, despite their utility in low-resource and/or low-quality scenarios.
\item[Scale.] Several datasets used in prior work were small with respect to the number of labels collected from crowd (\#labels). But on large-scale datasets, like \texttt{senti} used in our experiment with $569$k labels, the scalability of inference becomes critical. For example, GLAD~\citep{whitehill2009whose}, achieving the third highest mean accuracy among all existing methods in our experiment, takes more than $3$ hours to train, while MV runs in $3.5$ seconds.
\end{description}

This work is also motivated by qualitative considerations: ease of understanding, implementation, and hyper-parameter setting, for which MV is compelling, being parameter free. However, empirically the accuracy of MV is far from perfect, even for highly redundantly annotated datasets. We seek accurate, fast solutions to the high-resource high-quality scenario that are easy to implement. Our method is one such solution.

This paper presents a novel method for aggregating crowdsourced labels for classification.
We propose a generative Bayesian model, relaxing a discrete binary problem to a continuous regression task, which we model using a normal likelihood and conjugate inverse-Gamma prior.
Our method allows for straightforward inference using expectation-maximisation (EM; \cite{dempster1977maximum}), which is implemented in $50$ lines of code (see Appendix A).
The resulting update rules reveal a novel criterion for learning from crowds---namely the idea of minimizing the likelihood that ``every worker is making mistakes''.

Our base model has interpretable parameters and intuitive inference rules, which makes it easy to understand and use.
We extend our base binary model in three ways: estimating different weights for each label, to capture both the reliability of a worker votes in favour and against an item belonging to a class; a means of correcting for model deficiency; and employing the one-versus-rest strategy to allow the model to operate in a multi-class setting.
Our experimental results show our proposed methods achieve the highest mean accuracy over $10$ real-world datasets drawn from several different application areas, with different sizes and differing degrees of label redundancy, and a total running time on the same datasets second only to MV.

\section{Related work}\label{sec:related_work}
Ever since \citet{dawid1979maximum} proposed a model (DS) to aggregate the clinical diagnoses of doctors, various methods have been proposed to aggregate worker labels and infer true annotations.
The DS model uses the confusion matrix to capture the probabilities that a worker label is generated for an item conditioned on the item's true annotation.
Extending the DS model by adding Dirichlet priors for the values in the confusion matrix, 
\citet{kim2012bayesian} then integrated the confusion matrix parameters out and used Gibbs sampling for inference.
\citet{simpson2013dynamic} developed variational Bayesian inference for the model proposed in \citep{kim2012bayesian}, which is computationally efficient and outperforms the Gibbs sampling method. Furthermore, they analysed the clusters naturally formed by the inferred workers' confusion matrices.
\citet{venanzi2014community} assumed there are a fixed number of clusters of workers, with workers in the same cluster possessing similar confusion matrices instead of modelling individual confusion matrices as done in previous works. In this way worker clusters are formed at inference-time instead of post-inference.
Adding a Dirichlet process to the model to generate the Dirichlet priors on worker confusion matrices, \citet{moreno2015bayesian} also allowed workers to generate their own confusion matrices from the cluster's confusion matrix in order to increase model flexibility.

Unlike the above works, \citet{whitehill2009whose} didn't use confusion matrices, but borrowed ideas from Item Response Theory (IRT) and proposed an unsupervised version of previous IRT models. These estimate not only the true annotations and worker ability but also the difficulty of items. \citet{zhou2012learning} also developed an algorithm estimating the true annotations by using a minimax entropy principle, which assumes there are categorical distributions per worker-item pair and the worker labels are generated from those categorical distributions. The objective there is in finding the true annotations that minimize the maximum entropy of those categorical distributions. Such categorical distributions are more concentrated on worker labels and can be roughly considered as having higher likelihood of generating labels.

The database community has also independently studied the problem of resolving conflicts between information from different sources. \citet{demartini2012zencrowd} developed a model similar to the DS model that also used EM but just modelled a single accuracy parameter instead of a full confusion matrix. \citet{li2014confidence} adopted the Gaussian to model error between the truth and worker label, which is similar to our model but not a probabilistic model. They developed an iterative approach which alternatively estimated the truth based on variances, and set the variances to be the upper-bounds of their 95\% confidence interval based on the estimated truth. \citet{aydin2014crowdsourcing} adopted a two-step procedure that iteratively updates estimated truth and worker weights so that the sum of worker weight times the distance between worker label and estimated truth is minimized.

\section{Proposed models}

In this section, we first define the crowdsourced annotation aggregation problem, then describe our proposed model, Bayesian Weighted Average (BWA), for binary classification tasks, followed by an extension to multi-class classification tasks, and strategies to set its hyper-parameters.

\paragraph{Problem statement.} We assume that a task contains a number of homogeneous \emph{instances} or \emph{items}, where each item has its own \emph{true label}. The true labels are drawn from a \emph{label set}. All instances are uploaded to a crowdsourcing platform by requesters. Generally the requesters don't know the true labels, so they ask \emph{workers} to label them. Workers need to choose a label from the label set per item. Typically a worker only labels a small fraction of all items in a task. The labels collected from workers are called \emph{worker labels}. Finally, aggregation methods are used to infer the true annotations based on all worker labels.

\subsection{The Bayesian Weighted Average Model}\label{subsec:base_model}
We first consider binary classification tasks. Without loss of generality, let $y_{ij}\in\{0,1\}$ be worker $j$'s label to item $i$ and let $z_i\in\mathbb{R}$ represent item $i$'s true label. We model $z_i$'s as continuous variables to make tractable the following optimization problem for inferring the truth, $Z$,
\begin{align}
Z^* = \arg\max_Z P(Z|Y,\alpha),\label{eq:optimization_MAP}
\end{align}
where $Z$ and $Y$ are collections of their lowercase variables, $\alpha$ denotes all hyper-parameters, $P(Z|Y,\alpha)$ is the posterior distribution for $Z$, and $Z^*$ is the maximum a posteriori (MAP) estimate of $Z$. Further discussion of our modelling choices is provided in Section~\ref{sec:discussion}.

After $Z^*$ is computed, the discrete true label can be determined by rounding, i.e. the true label is $1$ if $z_i^*>0.5$, otherwise $0$. Frequently used symbols are defined in Table~\ref{tab:symbols}.

\begin{table}[t]
\centering
\caption{Mathematical notation.}\label{tab:symbols}
\begin{tabular}{@{}l@{\quad}l@{}}
\toprule
Symbol & Description \\\midrule
$N$ & number of items \\
$i$ & item index, $i\in\{1,\dots,N\}$ \\
$W$ &  number of workers \\
$j$ & worker index, $j\in\{1,\dots,W\}$ \\
$K$ & number of classes \\
$k,l$ & class/label index, $k,l\in\{0,\dots,K-1\}$ \\
$W_i$ & set of workers who have labelled item $i$ \\
$N_j$ & set of items that worker $j$ has labelled \\
\midrule
\multicolumn{2}{c}{Binary setting}\\
\midrule
$z_i$ & item $i$'s latent true label (continuous), $z_i\in\mathbb{R}$ \\
$y_{ij}$ & worker $j$'s observed label to item $i$ (discrete), $y_{ij}\in\{0,1\}$ \\
\midrule
\multicolumn{2}{c}{Multi-class setting}\\
\midrule
\multirow{2}{*}{$z_{ki}$}  & score given to item $i$ by model $k$ which classifies \\
& items into class-$k$ or not class-$k$, $z_{ki}\in\mathbb{R}$\\
\multirow{2}{*}{$y_{kij}$} & binary indicator, equalling $1$ if worker $j$ labels\\
& item $i$ as class $k$ otherwise $0$\\
\bottomrule
\end{tabular}
\end{table}

\subsubsection{The Generative Process and Joint Distribution}\label{subsubsec:base_model_generative_process}
\begin{figure}[t]
\centering
\begin{tikzpicture}
  \matrix[row sep=0.3cm, column sep=0.6cm] (DPC)
  {
    \node[latent] (v_j)  {$v_j$} ; &
    &
    \\
    \\
    \node[obs]    (y_ij) {$y_{ij}$} ; &
    &
    \node[latent] (z_i)  {$z_i$} ;
    \\
  };
  \node[const, above=of v_j, xshift=-0.5cm] (a_v)    {$a_v$}; %
  \node[const, above=of v_j, xshift=0.5cm]  (b_v)    {$b_v$}; %
  \node[const, above=of z_i]                (mu)     {$\mu$}; %
  \node[const, right=of z_i, xshift=-0.5cm] (lambda) {$\lambda$}; %

  \edge {a_v,b_v}     {v_j} ; %
  \edge {mu,lambda} {z_i} ; %
  \edge {v_j,z_i}   {y_ij} ; %

  \plate {plt_i} {
    (z_i)(y_ij)
  } {$i=1\dots N$} ;

  \plate {plt_j} {
    (v_j)(y_ij)(plt_i.south west)
  } {$j=1\dots W$} ;
\end{tikzpicture}
\caption{Plate diagram for our proposed BWA model (binary case).}\label{fig:pgm_base}
\end{figure}
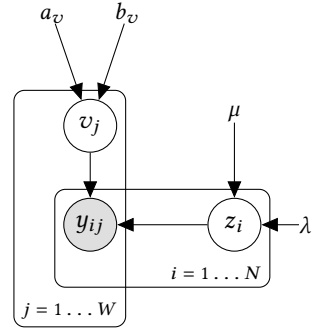

Our proposed models are all probabilistic generative models. We assume $z_i$ is drawn from a Gaussian prior with mean $\mu$ and inverse variance $\lambda$, and $y_{ij}$ is drawn from a Gaussian distribution with mean $z_i$ and inverse variance $v_j$. Here $v_j$ is a worker specific value for worker $j$ and every $v_j$ is drawn from $\mathrm{Gamma}(\frac{a_v}{2},\frac{b_v}{2})$. The plate diagram for this model is shown in Figure~\ref{fig:pgm_base}.

The joint distribution $P(Z,Y,V|\mu,\lambda,a_v,b_v)$ can be factorized as
\begin{align}
&P(Z,Y,V|\mu,\lambda,a_v,b_v) = P(Z|\mu,\lambda)\cdot P(Y|Z,V)\cdot P(V|a_v,b_v)\nonumber\\
=&\prod_i\mathcal{N}\left(z_i\middle|\mu,\lambda^{-1}\right)\nonumber\\
&\cdot\prod_i\prod_{j\in W_i}\mathcal{N}\left(y_{ij}\middle|z_i,v_j^{-1}\right)
\cdot\prod_j\mathrm{Gamma}\left(v_j\middle|\frac{a_v}{2},\frac{b_v}{2}\right),\label{eq:joint_base}
\end{align}
where $V = \{ v_j \}_{j=1}^W$. $\lambda$ and $v_j$ model the inverse variance of $z_i-\mu$ and $y_{ij}-z_i$. The larger $\lambda$ and $v_j$ are, the smaller the corresponding errors are likely to be. In this way they can be interpreted as encoding worker reliability, a perspective  we explore further in Section~\ref{subsubsec:base_model_interpretations}.

Our proposed BWA model is deficient: data observations $y_{ij}$ can only be binary, while the generative model has continuous support. We could force all $y_{ij}$ to be binary variables ($\{0,1\}$) then re-normalize the distributions accordingly to address the deficiency issue (Section~\ref{subsec:novel_constraint}). But the linking distribution between $y_{ij}$, $z_i$ and $v_j$ then becomes a logistic function, which is much more difficult to work with as no conjugate prior for $v_j$ exists.

\subsubsection{Inference}
In order to find the most likely $Z$, we first integrate out $V$ in Equation~\eqref{eq:joint_base} to obtain $P(Z,Y|\alpha)$ where $\alpha=(\mu,\lambda,a_v,b_v)$ are the hyper-parameters.
\begin{align*}
& P(Z,Y|\alpha) = \int P(Z,Y,V|\mu,\lambda,a_v,b_v)\;\mathrm{d}V\\
=&\prod_i\mathcal{N}(z_i|\mu,\lambda^{-1})\\
&\cdot\prod_j\int\prod_{i\in N_j}\mathcal{N}\left(y_{ij}\middle|z_i,v_j^{-1}\right)
\cdot\mathrm{Gamma}\left(v_j\middle|\frac{a_v}{2},\frac{b_v}{2}\right)\;\mathrm{d}v_j\\
\propto&\prod_i e^{-\frac{\lambda}{2}(z_i-\mu)^2}
\cdot\prod_j\left(\frac{1}{2}b_v+\frac{1}{2}\sum_{i\in N_j}(z_i-y_{ij})^2\right)^{-\frac{a_v+|N_j|}{2}}.
\end{align*}
Since $P(Z,Y|\alpha)=P(Z|Y,\alpha)P(Y|\alpha)$ and $P(Y|\alpha)$ is independent from $Z$, maximising $P(Z,Y|\alpha)$ with respect to $Z$ is equivalent to solving the program~\eqref{eq:optimization_MAP}, i.e.
$$\arg\max_Z P(Z|Y,\alpha) = \arg\max_Z P(Z,Y|\alpha).$$

As we shortly show, $\lambda$, $a_v$, and $b_v$ have intuitive meanings, and accordingly can be set manually, while $\mu$ can be optimized easily with $Z$. Therefore, the goal becomes to maximise $P(Z,Y|\alpha)$ with respect to $Z$ and $\mu$, which is equivalent to minimising the negative log likelihood $-\log P(Z,Y|\alpha)$ with respect to $Z$ and $\mu$:
\begin{align}
\min_{Z,\mu}&\sum_i\frac{\lambda}{2}(z_i-\mu)^2 + \sum_j\frac{a_v+|N_j|}{2}\log\left(b_v+\sum_{i\in N_j}(z_i-y_{ij})^2\right)\nonumber\\
& + \mathit{const.},\label{eq:neg_log_likelihood_base}
\end{align}
where $\mathit{const.}$ is the sum of values independent from $Z$ and $\mu$, and thus can be ignored during optimisation. We then use the EM algorithm to minimize the above negative likelihood:\\
\emph{E step.} $v_j|Z,Y,a_v,b_v
\sim\mathrm{Gamma}\left(\frac{a_v+|N_j|}{2},\frac{b_v+\sum_{i\in N_j}(z_i-y_{ij})^2}{2}\right)$.\\
\emph{M step.} Let $q(V)=P(V|Z^{\mathrm{old}},Y,a_v,b_v)$ and
$$Q(Z;Z^{\mathrm{old}}) = \mathbb{E}_q\log P(Z,Y,V|\mu,\lambda,a_v,b_v)\;,$$
then by setting the gradient of the $Q$ function with respect to $Z$ and $\mu$ to be zero, we obtain
\begin{align}
z_i &=\frac{\lambda\mu+\sum_{j\in W_i}\mathbb{E}_qv_jy_{ij}}{\lambda+\sum_{j\in W_i}\mathbb{E}_qv_j}\ ,
\label{eq:em_updates_base_z}\\
\intertext{where}
\mathbb{E}_qv_j &= \frac{a_v+|N_j|}{b_v+\sum_{i\in N_j}(z_i-y_{ij})^2}\ ,\label{eq:em_updates_base_v}\\
\mu &=\frac{1}{N}\sum_iz_i\ .\nonumber
\end{align}

\subsubsection{Interpretations}\label{subsubsec:base_model_interpretations}
\paragraph{$\mathbf{z_i}$ and its update rule.} In Equation~\eqref{eq:em_updates_base_z}, $z_i$ is updated by a weighted arithmetic average of $\mu$ and $\{y_{ij}\}_{j\in W_i}$, weighted by $\lambda$ and $\{\mathbb{E}_qv_j\}_{j\in W_i}$, respectively. Higher weights lead to a larger impact on average, so these weights can be considered as encoding worker reliability. In particular, $\lambda$ is the weight for a ``default worker'' who always labels every item as $\mu$.

\paragraph{$\mathbf{v_j}$ and its estimation rule.} In Equation~\eqref{eq:em_updates_base_v}$, |N_j|$ is the number of items that worker $j$ has labelled, $\sum_{i\in N_j}(z_i-y_{ij})^2$ is the sum of squared errors ($\mathrm{SSE}_j$) that worker $j$ has made. Note that $y_{ij}\in\{0,1\}$ and $z_i\in[0,1]$, thus $0\le\mathrm{SSE}_j\le|N_j|$. In the case that we know the true label, $z_i$ is also binary just like $y_{ij}$, then $\mathrm{SSE}_j$ is simply the number of wrong judgements that worker $j$ has made. The estimation of $\mathbb{E}_qv_j$ can be considered as a smoothed inverse error rate with $a_v$ and $b_v$ serving as the smoothing parameters. The larger the $\mathbb{E}_qv_j$, the smaller the worker $j$'s smoothed error rate is.

\paragraph{Hyper-parameters.} The values of $a_v$ and $b_v$ encode our belief that worker $j$ has labelled $a_v$ items before and $b_v$ judgements were wrong, which naturally suggests $b_v\le a_v$ and further ensures $\mathbb{E}_qv_j\ge 1$. So when $b_v\le a_v$, even if all labels from worker $j$ in task $k$ are wrong, we still have $\mathbb{E}_qv_j\ge 1$. This mechanism guarantees everyone an unconditional minimum weight of $1$ to vote for the truth. We may consider setting $b_v>a_v$ so that $\mathbb{E}_qv_j$ is allowed to be smaller than $1$, however, in this setting, a worker who gets $10/10$ wrong will be penalized more than another worker who gets $1000/1000$ wrong due to the weaker smoothing effect on the latter. Hence both estimations are smaller than $1$, but the estimation for the latter is closer to $1$.
As this is not the desired effect, we recommend the natural setting of $b_v\le a_v$. As $1$ is the lower bound of $\mathbb{E}_qv_j$, we choose $\lambda=1$ so that the ``default worker'' has limited ability to compete with other workers, but is most helpful in tie breaking.

\subsection{Multi-class Extension}
So far, our proposed models assume binary labels.
However, it's easy to extend them to the multi-class setting by employing a one-versus-rest strategy, a commonly used reduction to binary classifiers such as SVMs.

Let $K$ denote the number of classes, $k\in\{0,\dots,K\}$ a class index, $y_{kij}$ a binary indicator that equals $1$ if worker $j$ labels item $i$ as class $k$ otherwise $0$, $Y_k$ the collection of $y_{kij}$'s. Then our models can work on $Y_k$ which is binary and output $Z_k$, a collection of $z_{ki}$'s, where $z_{ki}$ is the score that our model gives to item $i$. The higher $z_{ki}$ is, the more likely our model believes item $i$'s true label is $k$. For item $i$, our model generates a score $z_{ki}$ for each $k$ value given different $Y_k$. Finally, $k^*=\arg\max_k z_{ki}$ is picked as the estimated true label for item $i$.



\subsection{Hyper-parameter Settings}\label{subsec:poe:set_bv}
We discuss how to set $b_v$ when $a_v$ is given in this section. As $a_v$ and $b_v$ translate to our prior belief
that every worker has labelled $a_v$ items and made $b_v$ mistakes, $b_v/a_v=\varepsilon$ is our assumed error rate.
This error rate should depend on the quality of worker labels in a dataset, so we seek to estimate $\varepsilon$
instead of setting it to be constant regardless of datasets. However, estimating $\varepsilon$ requires true labels
which we don't have, so a natural way is to estimate them by a simple aggregation method such as majority voting,
then calculate $\varepsilon$ based on the estimates.

We first consider the binary case. Let $\hat{z}_i=\frac{1}{|W_i|}\sum_{j\in W_i}y_{ij}$
to be the majority voting results, then the overall error rate $\varepsilon$ is
$$\varepsilon = \frac{\sum_i\sum_{j\in W_i}(y_{ij}-\hat{z}_i)^2}{\sum_i|W_i|}\ .$$
To simplify $\varepsilon$, we define $n_{i0}$ and $n_{i1}$ to be the number of workers
who have labelled item $i$ as $0$ or $1$, respectively.
Then $|W_i|=n_{i0}+n_{i1}$, $\hat{z}_i=\frac{n_{i1}}{n_{i0}+n_{i1}}$ and we obtain
\begin{align}
\varepsilon =& \frac{\sum_i n_{i0}(0-\hat{z}_i)^2 + n_{i1}(1-\hat{z}_i)^2}{\sum_i n_{i0}+n_{i1}}\nonumber\\
=& \frac{\sum_i n_{i0}(\frac{n_{i1}}{n_{i0}+n_{i1}})^2 + n_{i1}(\frac{n_{i0}}{n_{i0}+n_{i1}})^2}{\sum_i n_{i0}+n_{i1}}
= \frac{\sum_i\frac{n_{i0}n_{i1}}{n_{i0}+n_{i1}}}{\sum_i n_{i0}+n_{i1}}\ .\label{eq:poe:error_rate}
\end{align}
Then we can set $b_v=a_v\varepsilon$ to encode that every worker has already annotated $a_v$ items,
with $a_v\varepsilon$ mistakes, i.e. an error rate $\varepsilon$.

However, we find that the error rate is underestimated by Equation~\eqref{eq:poe:error_rate} in practice.
We then derive that the error rate given by Equation~\eqref{eq:poe:error_rate} has an upper bound $1/4$:
\begin{align*}
\varepsilon =& \frac{\sum_i\frac{1}{n_{i0}+n_{i1}}n_{i0}n_{i1}}{\sum_i n_{i0}+n_{i1}}\\
\le& \frac{\sum_i\frac{1}{n_{i0}+n_{i1}}\left(\frac{n_{i0}+n_{i1}}{2}\right)^2}{\sum_i n_{i0}+n_{i1}}
= \frac{\sum_i\frac{n_{i0}+n_{i1}}{4}}{\sum_i n_{i0}+n_{i1}} = \frac{1}{4}\ .
\end{align*}
So given $\varepsilon\in [0,1/4]$, setting $b_v=a_v\varepsilon$ means our prior belief is
that every worker has an accuracy of $75\%$ or higher even in the worst case, which is too optimistic
and often doesn't reflect reality. We then propose doubling $\varepsilon$ to extend its range to $[0,1/2]$
to encode our belief that every worker is performing slightly better than random guessing.

For multi-class tasks, as there are $K$ classes, we expect random guessing can achieve an accuracy of $1/K$,
i.e. an error rate of $1-\frac{1}{K}$. To encode that the prior belief that every worker is better than
random guessing, we extend the range of $\varepsilon$ to $[0,1-\frac{1}{K}]$ by multiplying $\varepsilon$
with $4(1-\frac{1}{K})$. This is consistent with the doubling strategy for binary tasks,
as $4(1-\frac{1}{K})=2$ when $K=2$.

\section{Discussion}\label{sec:discussion}
The primary motivation of modelling the truth $Z$ as continuous variables is to make the optimization of finding the most likely $Z$ tractable, i.e. $\arg\max_Z P(Z|Y,\alpha)$. We first describe the difference between this objective and other probabilistic models' objectives, then discuss some interesting findings due to the continuous modelling of $Z$.

\subsection{An Objective Intractable for Other Models}
Consider a graphical model defining joint $P(Z,Y,V|\alpha)$. It is often easy to sum out $Z$ or integrate out $V$ in $P(Z,Y,V|\alpha)$ to get $P(Y,V|\alpha)$ or $P(Z,Y|\alpha)$. Ideally, the most likely $Z$ is computed by
$$Z^*=\arg\max_Z P(Z|Y,\alpha)\textrm{, or equivalently }Z^*=\arg\max_Z P(Z,Y|\alpha),$$
because $P(Z,Y|\alpha)\propto P(Z|Y,\alpha)$. However, when $Z$ is discrete, as in most existing probabilistic models for truth inference, the above objective becomes an intractable discrete optimization task. Typical workarounds are:
\begin{enumerate}
\item
Solve $V^* = \arg\max_V P(Y,V|\alpha)$ first, then set $V=V^*$ into the joint distribution which naturally decouples $z_{i}$'s, so $Z^* = \arg\max_Z P(Y,Z,V^*|\alpha)$ becomes tractable as $z_i$'s can be optimized independently.
$$Z^* = \arg\max_Z P(Y,Z,V^*|\alpha)$$
where $V^* = \arg\max_V P(Y,V|\alpha).$
The compromise here is that although $V^*$ is optimal, $Z^*$ is only optimal when $V=V^*$, which does not capture uncertainty. Examples of methods taking this approach include DS~\citep{dawid1979maximum} and LFC~\citep{raykar2010learning}.
\item Bayesian methods naturally use the posterior to make inference. In $P(Z,Y,V|\alpha)$, only $Z$ is the variable of interest, so $V$ is marginalized first, leaving $P(Z,Y|\alpha)$. The posterior of $Z$, $P(Z|Y,\alpha)$ is proportional to $P(Z,Y|\alpha)$, however, the normalizing constant $P(Y|\alpha)$ is typically intractable to compute. There are two common ways to deal with this intractability.

\begin{itemize}
\item\emph{Variational approaches} \citep{jordan1999introduction} approximate the posterior \linebreak[4] $P(Z|Y,\alpha)$ by another distribution $q(Z)$, which is then used to make inference. Mean-field variational inference is very popular and assumes $q(Z)$ is factorized into $\prod_i q(z_i)$, then the true label of item $i$ is estimated as $\arg\max_{z_i}q(z_i)$, which approximates $\arg\max_{z_i}P(z_i|Y,\alpha)$ where $P(z_i|Y,\alpha)$ is the marginal distribution of $z_i$.

\item\emph{Sampling techniques} generate samples from $P(Z|Y,\alpha)$. \linebreak[4] Markov chain Monte Carlo methods are commonly used to generate samples as they can work without knowing the normalizing constant $P(Y|\alpha)$. Then samples are used to approximate the marginal distribution over $z_i$, i.e. $P(z_i|Y,\alpha)$. Finally, $\arg\max_{z_i}P(z_i|Y,\alpha)$ is picked as the estimate of item $i$'s true label.
\end{itemize}

In summary, with either variational approaches or sampling, the goal of Bayesian methods is to solve
$$z_i^* = \arg\max_{z_i}P(z_i|Y,\alpha).$$
However, both approaches face the shortcoming that the variables of interest ($z_{i}$) must be optimized independently according to their marginal distributions, as joint optimization is intractable as $Z$ is discrete. Examples include iBCC~\citep{kim2012bayesian} and CBCC~\citep{venanzi2014community}.
\end{enumerate}

\noindent
By contrast, our methods' objectives are ideal:
$$Z^* = \arg\max_Z P(Z|Y,\alpha).$$
The only compromise we make is that we have to model $Z$ as continuous random variables.

\subsection{A Novel Perspective on the Wisdom of Crowds}
For our proposed BWA model, the negative log likelihood \linebreak[4] $-\log P(Z,Y|\alpha)$ defined in Equation~\eqref{eq:neg_log_likelihood_base} is reminiscent of a traditional sum-of-square-error objective as minimized in regression models, except that here the error terms are contained in separate $\log$ functions. Reorganizing the expression, and using the sum of squared errors defined as $\mathrm{SSE}_j=\sum_{i\in N_j}(z_i-y_{ij})^2$, we obtain
\begin{align*}
-\log P&(Z,Y|\mu,\lambda,a_v,b_v) \\
=& \frac{\lambda}{2}\sum_i(z_i-\mu)^2 + \frac{1}{2}\sum_j\log\left(\frac{b_v+\mathrm{SSE}_j}{a_v+|N_j|}\right)^{a_v+|N_j|} + \mathit{const.}
\end{align*}
The exponential of the RHS yields the equivalent objective
\begin{align*}
\arg\min_{Z,\mu}-&\log P(Z,Y|\mu,\lambda,a_v,b_v)\\
=& \arg\min_{Z,\mu}\prod_{\textrm{item }i}e^{\lambda(z_i-\mu)^2}
\prod_{\textrm{worker }j}\left(\frac{b_v+\mathrm{SSE}_j}{a_v+|N_j|}\right)^{a_v+|N_j|}.
\end{align*}
The first product over items serves as a regularization term for $z_i$. For $\mu$, it simply means the minimum can be reached when $\mu$ is the mean of all $z_i$. The product over workers is more interesting. Note that $\frac{b_v+\mathrm{SSE}_j}{a_v+|N_j|}$ also appears in $\mathbb{E}_qv_j$'s update rule, and can be considered as the smoothed error rate of worker $j$.
The exponent, $a_v+|N_j|$, is just the smoothed number of items worker $j$ has labelled.
Accordingly, this factor measures the likelihood that all workers make all their annotations of the data incorrectly, with a uniform error rate per worker.

Therefore, the underlying principle behind our proposed BWA model is that \emph{it's unlikely that all workers make mistakes for all their tasks}. Our proposed BWA model infers a truth that minimizes the likelihood of everybody performing incorrectly. This principle makes our model different from other models
which seek to explain worker labels through maximising the likelihood of realizing these observations.
Our principle is conservative, yet intuitively plausible, and a novel perspective on the \emph{wisdom of crowds}.


\subsection{A Novel Constraint on Worker Reliability}\label{subsec:novel_constraint}
In the base model, if we were to force all $y_{ij}$ and $z_i$ to be binary variables ($\{0,1\}$) and then re-normalize the distributions accordingly, we obtain a new link function between $y_{ij}$ and $z_i$,
\begin{align*}
P(y_{ij}=1|z_i,v_j) &= \frac{\sqrt{\frac{v_j}{2\pi}}\exp\left\{-\frac{v_j}{2}(z_i-1)^2\right\}}
{\sqrt{\frac{v_j}{2\pi}}\exp\left\{-\frac{v_j}{2}z_i^2\right\}+\sqrt{\frac{v_j}{2\pi}}\exp\left\{-\frac{v_j}{2}(z_i-1)^2\right\}}\\
&=\frac{1}{\exp\left\{-v_j(z_i-\frac{1}{2})\right\}+1}\textrm{ for }z_i\in\{0,1\},
\end{align*}
or in an equivalent but more compact form,
\begin{align*}
P(y_{ij}|z_i,v_j) = \frac{\exp\left\{\frac{1}{2}v_j\mathbf{1}[y_{ij}=z_i]\right\}}{1+\exp\left\{\frac{1}{2}v_j\right\}},
\end{align*}
which means each worker has an accuracy of $\sqrt{e^{v_j}}/(1 + \sqrt{e^{v_j}})$, irrespective of the true label $z_i$.
Such behaviour can be captured by a symmetric confusion matrix
$$\frac{1}{1 + \sqrt{e^{v_j}}} \begin{bmatrix} \sqrt{e^{v_j}} & 1 \\ 1 & \sqrt{e^{v_j}} \end{bmatrix}.$$

Due to $v_j$ having a Gamma prior, $v_j$ must be non-negative. Consequently the values on the diagonal must be larger than or equal to the values off diagonal.
Accordingly our approach is capable of modelling:
a perfect worker,
 $\left[\begin{smallmatrix}1&0\\0&1\end{smallmatrix}\right]$,
a typical worker,
$\left[\begin{smallmatrix}0.7&0.3\\0.3&0.7\end{smallmatrix}\right]$,
and a random worker,
$\left[\begin{smallmatrix}0.5&0.5\\0.5&0.5\end{smallmatrix}\right]$.
However adversarial workers, e.g., $\left[\begin{smallmatrix}0.2&0.8\\0.8&0.2\end{smallmatrix}\right]$, cannot be represented.
This limitation arises from the choice of prior (Gamma), and does not feature in other confusion matrix based models.

This constraint means our models can't capture adversarial workers, however this limitation fits with empirical observations of crowd-workers: in practice adversarial workers are rare, and are largely filtered out based on performance history, which is a consequence of the data acquisition method, but it's possible that less carefully collected data with less rigorous quality control may not be a good fit to our model. Typically in a confusion matrix based probabilistic model, worker labels are assumed to be drawn from categorical distributions parameterized by rows in confusion matrices, and every row in a confusion matrix is assumed to be drawn from a Beta or Dirichlet distribution. For binary tasks, the constraint that values on the diagonal are larger than or equal to the values off the diagonal could be achieved by using a truncated Beta prior with a support $[0.5, 1]$, but at a much higher computational cost because the distributions will no longer be conjugate.


\section{Experiments}\label{sec:experiment}
\paragraph{Initialization.}
Our techniques use the EM algorithm, and thus require initialization of either $\mu$ and $\mathbb{E}_qV$, or $Z$. We initialise $Z$, the easier of the two, using majority voting which allows for fast convergence. In some cases, from the MV initialization our algorithm doesn't converge to the global optimum (the highest likelihood).
A common solution is to run the algorithm with random initialization multiple times and then find the best likelihood (including a majority voting run),
but to obtain deterministic results, we only run our algorithm with the MV initialization in all experiments.

Since we use the released implementations of existing methods, we choose to keep their code unchanged. Implementations initializing parameters randomly (CATD, CRH) are run 10 times on every dataset and the mean accuracy of 10 runs is reported as its accuracy on that dataset.

\paragraph{Hyperparameter settings.} Following the suggestions described in
Section~\ref{subsubsec:base_model_interpretations}, we set $\lambda=1$.
We choose $a_v$ and $b_v$ in Section~\ref{subsec:poe:choose_av}.
We set the tolerance in the stopping criteria to $10^{-3}$, such that iteration of the algorithm stops
when the relative difference between every $z_i$ and its last value is within $0.1\%$.

\paragraph{Methods.}
\citet{zheng2017truth}\footnote{https://zhydhkcws.github.io/crowd\_truth\_inference/index.html}
compared 17 existing aggregation methods and released their implementations,
and 10 of them supporting the multi-class setting are used in our experiment
including MV, ZC~\citep{demartini2012zencrowd}, GLAD~\citep{whitehill2009whose},
DS~\citep{dawid1979maximum}, Minimax~\citep{zhou2012learning}, iBCC~\citep{kim2012bayesian},
CBCC~\citep{venanzi2014community}, LFC~\citep{raykar2010learning}, CATD~\citep{li2014confidence},
and CRH~\citep{aydin2014crowdsourcing}.

\paragraph{Datasets.}
There are $19$ real-world datasets used in the experiments originating with four crowdsourcing dataset collections,
namely the union of
the CrowdScale 2013's shared task challenge~\citep{josephy2014workshops}%
\footnote{https://sites.google.com/site/crowdscale2013/shared-task} (2 datasets),
the Active Crowd Toolkit project~\citep{venanzi2015activecrowdtoolkit}%
\footnote{https://github.com/orchidproject/active-crowd-toolkit} (8 datasets),
the Truth Inference in Crowdsourcing project~\citep{zheng2017truth}%
\footnote{See footnote 1.} (7 datasets),
and the GitHub repository for SpectralMethodsMeetEM paper~\citep{zhang2014spectral}%
\footnote{https://github.com/zhangyuc/SpectralMethodsMeetEM} (5 datasets).
Note that $3$ datasets are in common between the last two collections.
Table~\ref{tab:datasets} presents some statistics about the 19 datasets.

\begin{table*}[t]
\caption{Statistics about 19 real-world datasets used in our experiments.}
\label{tab:datasets}
\vskip -0.05in
\centering
\begin{tabular}{>{\tt}ld{0}d{0}d{0}d{0}d{0}d{2}d{2}}
\toprule
Dataset & \mathrm{\#item}(N) & \mathrm{\#worker}(W) & \mathrm{\#class} & \mathrm{\#label} & \mathrm{\#truth}
        & \multicolumn{1}{c}{$\mathrm{\#label}/N$} & \multicolumn{1}{c}{$\mathrm{\#label}/W$} \\ \midrule
senti   & 98980  & 1960     & 5       & 569282  & 1000    & 5.75           & 290.45           \\
fact    & 42624  & 57       & 3       & 214960  & 576     & 5.04           & 3771.23          \\ \midrule
CF      & 300    & 461      & 5       & 1720    & 300     & 5.73           & 3.73             \\
CF*     & 300    & 110      & 5       & 6025    & 300     & 20.08          & 54.77            \\
MS      & 700    & 44       & 10      & 2945    & 700     & 4.21           & 66.93            \\
SP      & 4999   & 203      & 2       & 27746   & 4999    & 5.55           & 136.68           \\
SP*     & 500    & 143      & 2       & 10000   & 500     & 20             & 69.93            \\
ZCall   & 2040   & 78       & 2       & 20372   & 2040    & 9.99           & 261.18           \\
ZCin    & 2040   & 25       & 2       & 10626   & 2040    & 5.21           & 425.04           \\
ZCus    & 2040   & 74       & 2       & 11271   & 2040    & 5.53           & 152.31           \\ \midrule
prod    & 8315   & 176      & 2       & 24945   & 8315    & 3              & 141.73           \\
tweet   & 1000   & 85       & 2       & 20000   & 1000    & 20             & 235.29           \\
dog     & 807    & 109      & 4       & 8070    & 807     & 10             & 74.04            \\
face    & 584    & 27       & 4       & 5242    & 584     & 8.98           & 194.15           \\
adult   & 11040  & 825      & 4       & 89948   & 333     & 8.15           & 109.03           \\ \midrule
bird    & 108    & 39       & 2       & 4212    & 108     & 39             & 108              \\
trec    & 19033  & 762      & 2       & 88385   & 2275    & 4.64           & 115.99           \\
web     & 2665   & 177      & 5       & 15567   & 2653    & 5.84           & 87.95            \\
rte     & 800    & 164      & 2       & 8000    & 800     & 10             & 48.78            \\ \bottomrule
\end{tabular}
\vskip -0.1in
\end{table*}

The first two, \texttt{senti} and \texttt{fact}, are from the CrowdScale 2013's shared task challenge,
and are the largest two in our experiment, with 569k and 217k labels respectively.
The \texttt{senti} data asks the workers to judge the sentiment of a tweet discussing weather
and label it within \{Negative, Neutral, Positive, Unrelated, Can't tell\}.
The \texttt{fact} data requests workers to judge whether relational statements about public figures are correct,
e.g. ``Stephen Hawking graduated from Oxford''. Workers may choose from \{Yes, No, Skip\}.
Note that ``Can't tell'' or ``Skip'' may also appear as the true labels of some items due to
the inherent ambiguity of some questions. None of the methods compared in our experiments explicitly model
the unsure options, instead treating these as a standard label in a $5$ and $3$-class setting, respectively.

The next 8 datasets are all from the Active Crowd Toolkit project.
\texttt{MS}, collected by~\citet{rodrigues2013learning}, asks workers to listen to short music samples
and to classify the music into 10 music genres,
\{country, jazz, disco, pop, reggae, rock, blues, classical, mental, hip-hop\}.
\texttt{ZC\_in} and \texttt{ZC\_us}, provided by~\citet{demartini2012zencrowd},
ask workers to judge if a provided Uniform Resource Identifier (URI) is relevant to a named entity extracted from news,
where every URI describes an entity. \texttt{ZC\_all} is a combination of \texttt{ZC\_in} and \texttt{ZC\_us},
and all \texttt{ZC} datasets share the same set of items.
The two \texttt{SP}s are movie review datasets with binary sentiment labels.
\texttt{CF} is just a small part of the full \texttt{senti}.
\texttt{CF\textsuperscript{*}} and \texttt{SP\textsuperscript{*}} have the same items as \texttt{CF} and \texttt{SP},
but were reannotated.

The next 5 datasets are from the Truth Inference project~\citet{zheng2017truth}.
\texttt{prod}, collected by~\citet{wang2012crowder} for entity resolution,
asks workers whether two product descriptions refer to the same product or not.
\texttt{tweet}, collected by~\citet{zheng2017truth}, contains tweets where every tweet is related to a company.
Workers are asked to identify if the tweet has positive sentiment towards the company in the tweet.
\texttt{dog} contains dog images from ImageNet~\citep{krizhevsky2012imagenet} and the task is
to recognize a breed (out of Norfolk Terrier, Norwich Terrier, Irish Wolfhound, and Scottish Deerhound) for a given dog.
\texttt{face}, provided by~\citet{mozafari2014scaling}, contains face images,
and workers are asked to judge whether the person in the image is happy, sad, angry, or neutral.
\texttt{adult}, released by~\citet{mason2012conducting}, asks workers to determine
the age appropriateness (P, PG, R, X) of a website given its link.

The last five datasets are used in~\citet{zhang2014spectral}.
\texttt{bird}, provided by~\citet{welinder2010multidimensional}, is a dataset of bird images.
The task is to determine whether an image contains at least one duck.
The remaining three cover: recognising textual entailment, \texttt{rte}~\cite{snow2008cheap},
assessing the quality of retrieved documents, \texttt{trec} (TREC
2011 crowdsourcing track)\footnote{https://sites.google.com/site/treccrowd/2011},
judging the relevance of web search results, \texttt{web}~\cite{zhou2012learning}.

\subsection{Choosing $a_v$ and $b_v$}\label{subsec:poe:choose_av}
\begin{figure}[t]
  \centering
  \includegraphics[width=0.49\textwidth]{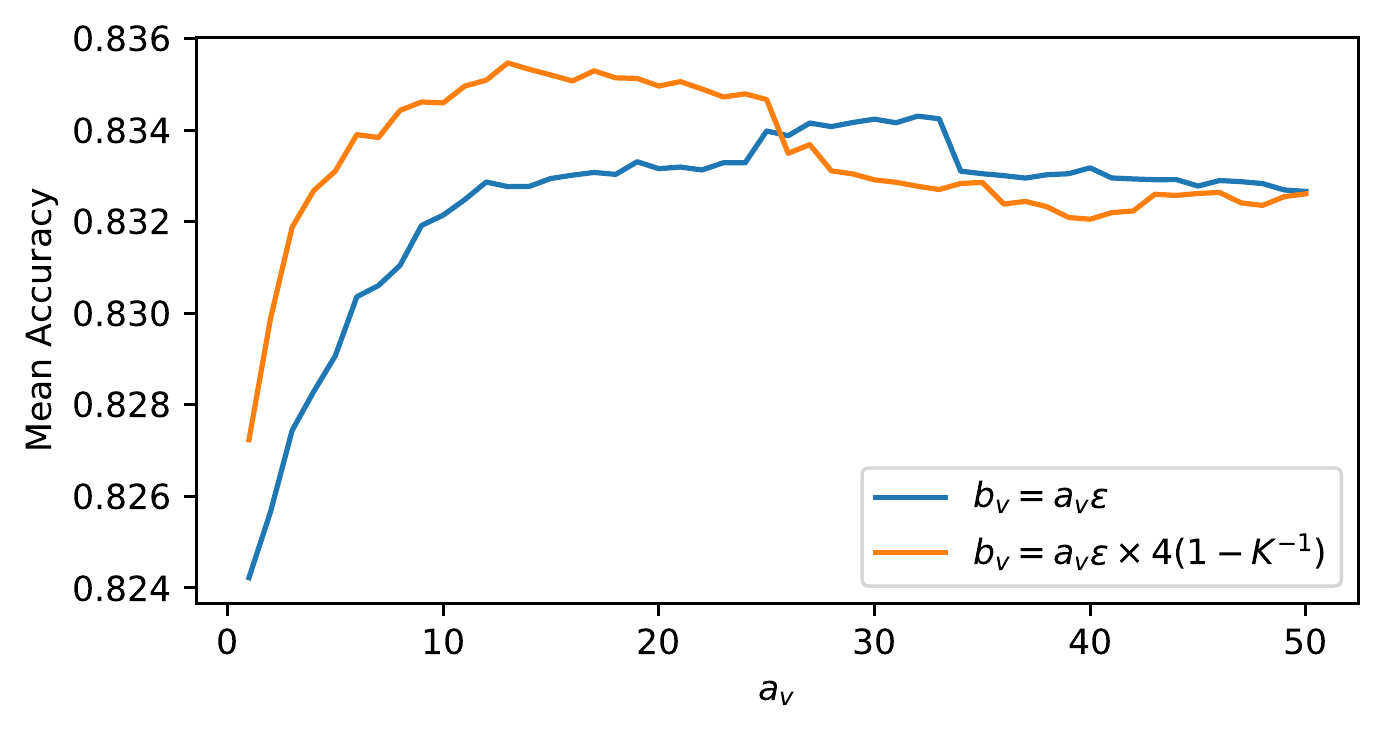}
  \vskip -0.1in
  \caption{The performance of our proposed BWA model with $b_v$ set by two different strategies.}
  \label{fig:poe:sensitivity}
  \vskip -0.15in
\end{figure}

Figure~\ref{fig:poe:sensitivity} shows the mean accuracy of our proposed BWA model across 19 datasets
with $b_v$ set by two different strategies described in Section~\ref{subsec:poe:set_bv}, namely,
using the original error rate $\varepsilon$ and using the adjusted error rate $\varepsilon\cdot4(1-\frac{1}{K})$.
For both strategies, the performance increases steeply when $a_v<10$ as the prior becomes stronger and provides
regularisation to workers who have labelled very few items. We also see both curves are flat around the optimum,
so the performance is not very sensitive to the choice of $a_v$.
Due to this reason, we pick $a_v$ from the flat region and report results in both settings:
BWA($a_v=30,\textrm{original }\varepsilon$) and BWA($a_v=15,\textrm{adjusted }\varepsilon$).

\subsection{Comparison against MV}
In this section, we report the results of one-sided Wilcoxon signed rank test performed on every method against MV.
We adopt the original version which discards the observations when two methods achieve the same accuracy on a dataset.
The null hypothesis $H_0$ is that the performance difference between a method and MV follows a symmetric distribution around zero.

Table~\ref{tab:wilcoxon} summarizes the results showing whether every method is significantly better than MV and at which level,
where significance levels
are denoted by asterisks, i.e. $0.1$(*), $0.05$(**), $0.025$(***), respectively.
$N_r$ denotes the effective sample size, and $W_-$ the statistic of a method summing up the ranks of datasets that this method is inferior to MV.
Zencrowd and iBCC achieve $85.33\%$ accuracy on the \texttt{CF\textsuperscript{*}} dataset
which is exactly the same performance as MV resulting,
therefore 
their $N_r$'s are reduced to $18$.

\begin{table}[tp]
\caption{One-sided Wilcoxon signed rank test results (all methods against MV)}
\label{tab:wilcoxon}
\vskip -0.05in
\centering
\resizebox{0.48\textwidth}{!}{%
\begin{tabular}{l>{$}c<{$}|>{$}c<{$}>{$}c<{$}|l|>{$}c<{$}}
\toprule
\multirow{2}{*}{Method} & \multicolumn{1}{c|}{Mean}     & \multirow{2}{*}{$N_r$} & \multirow{2}{*}{$W_-$} & \multicolumn{1}{c|}{Sig.}  & \multicolumn{1}{c}{Approx.}  \\
                        & \multicolumn{1}{c|}{accuracy} &                        &                        &
\multicolumn{1}{c|}{level} & \multicolumn{1}{c}{p-value}  \\ \midrule
MV                                          & 0.8196 &    &    &     &        \\
ZC                                          & 0.8294 & 18 & 64 &     & 0.1746 \\
GLAD                                        & 0.8305 & 19 & 47 & **  & 0.0267 \\
DS                                          & 0.8309 & 19 & 71 &     & 0.1671 \\
Minimax                                     & 0.8141 & 19 & 81 &     & 0.2866 \\
iBCC                                        & 0.8352 & 18 & 52 & *   & 0.0723 \\
CBCC                                        & 0.8291 & 19 & 71 &     & 0.1671 \\
LFC                                         & 0.8338 & 19 & 69 &     & 0.1477 \\
CATD                                        & 0.8334 & 19 & 56 & *   & 0.0583 \\
CRH                                         & 0.8054 & 19 & 91 &     & 0.4361 \\ \midrule
BWA($a_v=30,\textrm{original }\varepsilon$) & 0.8342 & 18 & 38 & *** & 0.0193 \\
BWA($a_v=15,\textrm{adjusted }\varepsilon$) & 0.8352 & 18 & 39 & *** & 0.0214 \\
\bottomrule
\end{tabular}
}
\vskip -0.1in
\end{table}

The rightmost column in Table~\ref{tab:wilcoxon} shows the p-value calculated based on approximating
the cumulative density function (cdf) of $W_-$ given $H_0$ as a cdf of Gaussian. Due to the small $N_r$
in our experiment ($18$ or $19$), the approximation is somewhat inaccurate, but we keep them there as a reference,
as they show ranking of the improvement achieved by all methods comparing to MV.

\subsection{Discussion}
\paragraph{Scalability.}
To measure the efficiency of all methods, we run all experiments on a modest desktop with a $7$-th generation Intel i$5$ CPU and $8$ GB memory.
All methods including our proposed BWA model are implemented in Python, except for iBCC and CBCC which are implemented in \texttt{C\#}, and Minimax is implemented in \texttt{MATLAB}.

\begin{figure}[t]
  \centering
  \includegraphics[width=0.48\textwidth]{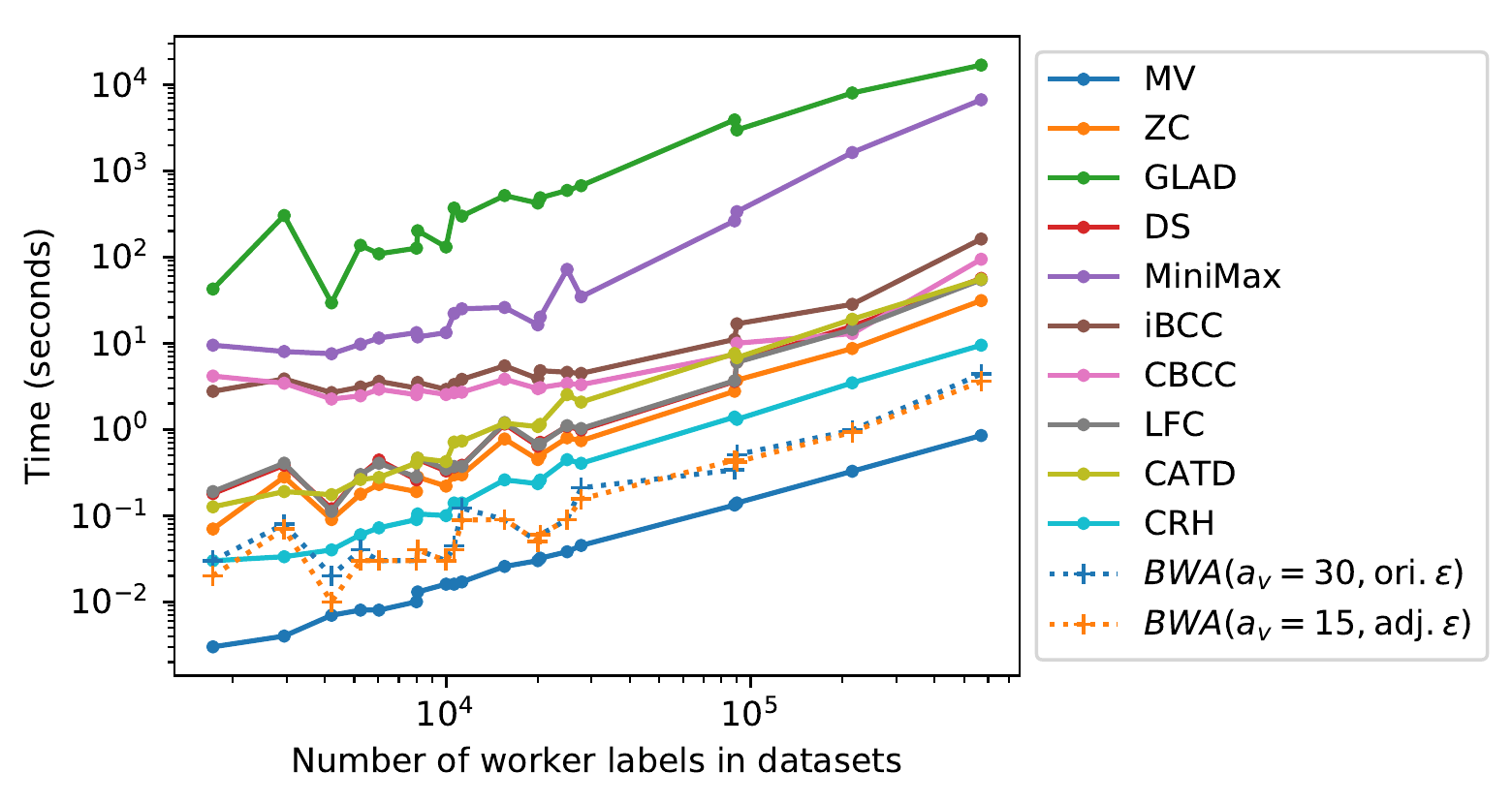}
  \vskip -0.1in
  \caption{Running time (seconds) of all methods on 19 real-world datasets.}
  \label{fig:scale}
  \vskip -0.13in
\end{figure}

As shown in Figure~\ref{fig:scale}, our proposed methods outperform the other existing methods (except MV) in terms of running time.
For datasets other than the largest two (\texttt{senti} and \texttt{fact}), our method converged within one second.
GLAD and Minimax are the two slowest methods, as they both require gradient-based optimization algorithms, and the required gradients are expensive to compute.

\paragraph{Accuracy}
As shown in Table~\ref{tab:wilcoxon}, although iBCC has the highest mean accuracy across all datasets,
it outperforms MV only at a significance level of $0.1$, which is insufficient
to reject the null hypothesis. On the other hand, GLAD, whose mean accuracy is not outstanding,
significantly outperforms MV at a significance level of $0.05$, the best among all 9 benchmarks.
In contrast, BWA achieves not only the highest mean accuracy, the same as iBCC, but also strongest significance of outperforming MV, at a level of $0.025$.

As shown in Table~\ref{tab:wilcoxon}, our proposed BWA($a_v=15$, adjusted $\varepsilon$) model and iBCC achieve the highest mean accuracy of $83.52\%$, and our BWA model using another hyper-parameter setting ($a_v=30$, original $\varepsilon$) which achieves $83.42\%$, followed by LFC $83.38\%$ and CATD $83.34\%$. Interestingly, both our proposed methods and CATD model the latent truth $z_i$'s as continuous variables, but in CATD the optimization objective is heuristic based, while in our models, the optimization objective is derived from explicit probabilistic models. Minimax is an interesting model in that it largely outperforms others on \texttt{bird} and \texttt{web}
but performs the worst on \texttt{MS}, the three \texttt{ZC} datasets, and \texttt{prod}.
This may be due to Minimax not being a probabilistic model thus its objective function is not well regularised
and often too aggressive. This may also explain the results for CRH, another non-probabilistic model.

\begin{figure*}[p]
  \centering
  \includegraphics[width=0.85\textwidth]{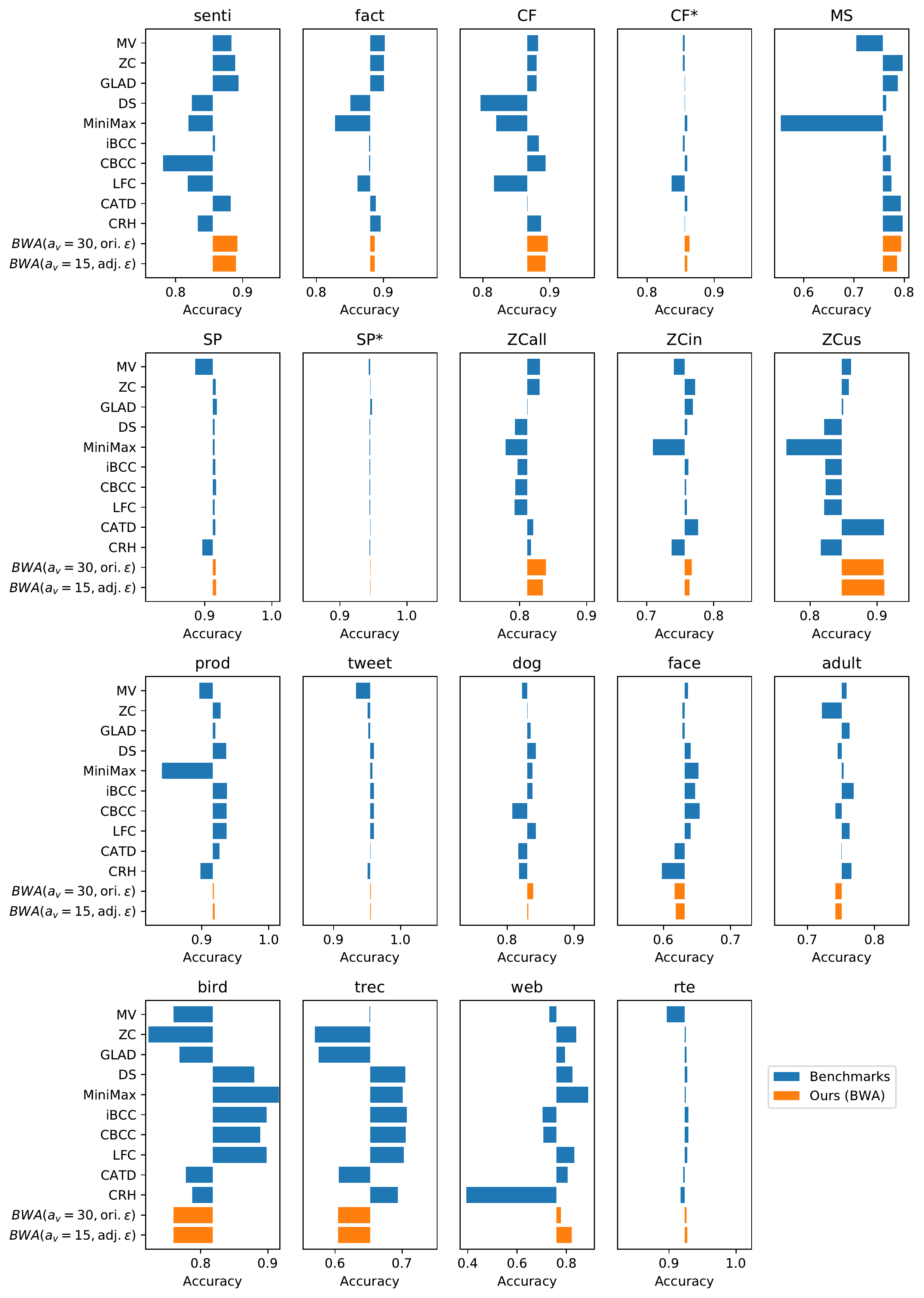}
  \caption{Accuracies of all methods on 19 real-world datasets. The horizontal baseline is the mean accuracy across all methods.}
  \label{fig:all_acc_real_data}
\end{figure*}

Figure~\ref{fig:all_acc_real_data} provides more details of the accuracies of all methods on the 19 real-world datasets.
We roughly sort all methods into two categories: confusion-matrix-based methods (DS, MiniMax, iBCC, CBCC, LFC) and 1-coin models (ZC, GLAD, CATD, CRH, our proposed BWA).
The name ``1-coin'' arises from these models only learning a single parameter per worker to capture their accuracy.
\texttt{face}, \texttt{bird}, \texttt{trec} are three successful cases for confusion-matrix-based methods,
because on those datasets workers behave very differently across classes, which ``1-coin'' models cannot capture.
However, they perform not so well on \texttt{senti}, \texttt{fact}, \texttt{MS}, and \texttt{ZC} datasets due to
either the data is overall insufficient to learn a $K\times K$ confusion matrix for every worker,
or the class is imbalanced that for some minority classes worker behaviour can't be reliably estimated.

\section{Conclusion}
This paper presents a novel method for aggregating crowdsourced labels for classification. We propose a generative Bayesian model, relaxing a discrete binary problem to a continuous regression task, which we model using a normal likelihood and conjugate inverse-Gamma prior, and can be extended to work in multi-class settings by employing the one-versus-rest strategy.

Our models allow for straightforward inference using expectation-maximisation, which is implemented in $50$ lines of code in the appendix. Our models have interpretable parameters and intuitive update rules, which make them easy to understand and use. We also provide detailed practical suggestions such as how to set the hyper-parameters.

Our experimental results show that our proposed method is the only one that significantly outperforms MV at significance level $0.025$ and performs competitively to the state-of-the-art method in terms of mean accuracy, across all datasets drawn from several different application areas, with different sizes and differing degrees of label redundancy. Our proposed method also has a total running time on the same datasets second only to MV.

\section*{Acknowledgement}
This work was sponsored by the Defense Advanced Research Projects Agency Information Innovation Office (I2O) under the Low Resource Languages for Emergent Incidents (LORELEI) program issued by DARPA/I2O under Contract No. HR0011-15-C-0114. The views expressed are those of the authors and do not reflect the official
policy or position of the Department of Defense or the U.S. Government. Trevor Cohn and Benjamin Rubinstien were supported by the Australian Research Council, FT130101105 and DP150103710, respectively.

\onecolumn
\appendix
\section{Code}
{\scriptsize
\begin{lstlisting}[language=Python]
import numpy as np
import pandas as pd
import scipy.sparse as ssp

def bwa_binary(y_exists_ij, y_is_one_ij, W_i, lambda_, a_v, adj_coef):
    N_j = y_exists_ij.sum(axis=0)
    z_i = y_is_one_ij.sum(axis=-1) / y_exists_ij.sum(axis=-1)
    
    b_v = a_v * W_i.dot(np.multiply(z_i, 1-z_i)) / y_exists_ij.sum() * adj_coef
    for _ in range(500):
        last_z_i = z_i.copy()
        
        mu  = z_i.mean()
        v_j = (a_v+N_j) / (b_v + (y_exists_ij.multiply(z_i)-y_is_one_ij).power(2).sum(0))
        z_i = (lambda_*mu + y_is_one_ij.dot(v_j.T)) / (lambda_ + y_exists_ij.dot(v_j.T))
        
        if np.allclose(last_z_i, z_i, rtol=1e-3): break
    return z_i.A1

def bwa(tuples, a_v=30, lambda_=1, prior_correction=True):
    num_items, num_workers, num_classes = tuples.max(axis=0) + 1
    num_labels = tuples.shape[0]
    W_i = np.bincount(tuples[:, 0])
    
    adj_coef = 4 * (1 - 1 / num_classes) if prior_correction else 1
    
    y_exists_ij = ssp.coo_matrix((np.ones(num_labels), tuples[:, :2].T),
                                 shape=(num_items, num_workers), dtype=np.bool).tocsr()
    y_is_one_kij = []
    for k in range(num_classes):
        selected = (tuples[:, 2] == k)
        y_is_one_kij.append(ssp.coo_matrix((np.ones(selected.sum()), tuples[selected, :2].T),
                                           shape=(num_items, num_workers), dtype=np.bool).tocsr())
    z_ik = np.empty((num_items, num_classes))
    for k in range(num_classes):
        z_ik[:, k] = bwa_binary(y_exists_ij, y_is_one_kij[k], W_i, lambda_, a_v, adj_coef)
    return z_ik

def get_acc(predictions, df_truth):
    score = (predictions == predictions.max(axis=1, keepdims=True)).astype(np.float)
    score /= score.sum(axis=1, keepdims=True)
    return score[df_truth.item.values, df_truth.truth.values].sum() / df_truth.shape[0]

data_path = '....'
datasets = ['crowdscale2013/sentiment', 'crowdscale2013/fact_eval']
records = []
for dataset in datasets:
    df_label = pd.read_csv(data_path + dataset + '/label.csv')
    df_label = df_label.drop_duplicates(keep='first')
    prediction_ik = bwa(df_label.values)

    df_truth = pd.read_csv(data_path + dataset + '/truth.csv')
    records.append((dataset, get_acc(prediction_ik, df_truth)))
print(pd.DataFrame.from_records(records, columns=['dataset', 'accuracy']))
## example output
#                     dataset  accuracy
# 0  crowdscale2013/sentiment  0.888000
# 1  crowdscale2013/fact_eval  0.885417

## label.csv example: | truth.csv example:
#                     |
#  item,worker,label  | item,truth
#  26440,1707,3       | 110,0
#  26440,311,3        | 161,1
#  26440,1356,3       | 328,1
#  26440,1774,3       | 611,2
#  ...                | ...
\end{lstlisting}
}

\twocolumn

\bibliographystyle{ACM-Reference-Format}
\balance
\bibliography{refs}

\end{document}